\documentclass[10pt,twocolumn,letterpaper]{article}

\usepackage{cvpr}              
\usepackage{lineno}
\newcommand\blfootnote[1]{%
  \begingroup
  \renewcommand\thefootnote{}\footnote{#1}%
  \addtocounter{footnote}{-1}%
  \endgroup
}
\definecolor{cvprblue}{rgb}{0.21,0.49,0.74}
\usepackage[pagebackref,breaklinks,colorlinks,allcolors=cvprblue]{hyperref}
\usepackage{graphicx}

\usepackage{booktabs}
\usepackage{multirow}
\usepackage{makecell}
\usepackage{colortbl}
\def\paperID{5254} 
\def\confName{CVPR}
\def\confYear{2025}

\title{DifIISR: A Diffusion Model with Gradient Guidance for \\
Infrared Image Super-Resolution}
\author{
  Xingyuan Li\textsuperscript{\rm 1}\thanks{Equal contribution.}, \quad Zirui Wang\textsuperscript{\rm 1}\footnotemark[1], \quad Yang Zou\textsuperscript{\rm 2}, \quad Zhixin Chen\textsuperscript{\rm 3},\\ Jun Ma\textsuperscript{\rm 1}, \quad Zhiying Jiang\textsuperscript{\rm 4}, \quad Long Ma\textsuperscript{\rm 1}, \quad Jinyuan Liu\textsuperscript{\rm 1}$^{\dagger}$\\
  \textsuperscript{1}Dalian University of Technology \hspace{0.1cm}
  \textsuperscript{2} Northwestern Polytechnical University \\
  \textsuperscript{3} Waseda University \hspace{0.1cm}
  \textsuperscript{4} Dalian Maritime University\\
  {\tt\small xingyuan\_lxy@163.com} \hspace{0.1cm}
  {\tt\small ziruiwang0625@gmail.com} \hspace{0.1cm}
}

\begin{document}
\twocolumn[{%
\renewcommand\twocolumn[1][]{#1}%
\maketitle

\begin{center}
    \centering
    \captionsetup{type=figure}
    \vspace{-0.15in}
    \includegraphics[width=1\textwidth]{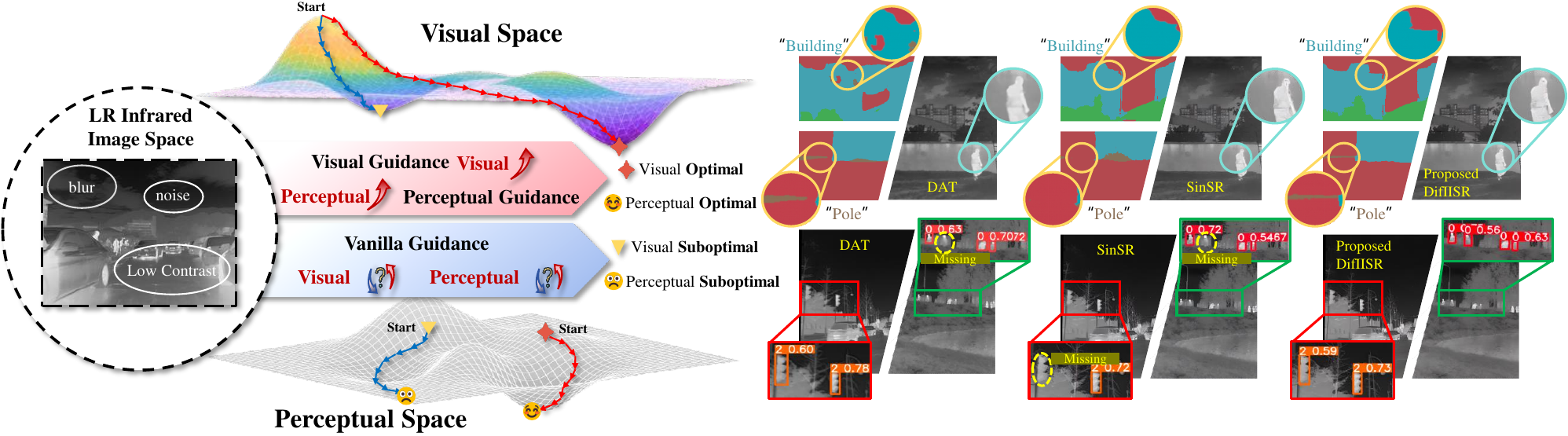}
    \vspace{-0.2in}
    \captionof{figure}{The left side shows a comparison between existing super-resolution methods and our proposed DifIISR. 
    Our method introduces additional visual guidance based on the Fourier Transform, as well as foundational model-based perception guidance. This allows our approach to achieve optimal performance in both visual and perceptual space. The right side demonstrates that our method outperforms other methods in both detection and segmentation tasks.}
    \label{fig:teaser}
\end{center}%

}]

\begin{abstract}

\blfootnote{$^*$ Equal contribution.  $^{\dagger}$ Corresponding author.}

Infrared imaging is essential for autonomous driving and robotic operations as a supportive modality due to its reliable performance in challenging environments. Despite its popularity, the limitations of infrared cameras, such as low spatial resolution and complex degradations, consistently challenge imaging quality and subsequent visual tasks. Hence, infrared image super-resolution (IISR) has been developed to address this challenge. While recent developments in diffusion models have greatly advanced this field, current methods to solve it either ignore the unique modal characteristics of infrared imaging or overlook the machine perception requirements. To bridge these gaps, we propose \textbf{DifIISR}, an infrared image super-resolution diffusion model optimized for visual quality and perceptual performance. Our approach achieves \textbf{task-based guidance} for diffusion by injecting gradients derived from visual and perceptual priors into the noise during the reverse process. Specifically, we introduce an infrared thermal spectrum distribution regulation to preserve visual fidelity, ensuring that the reconstructed infrared images closely align with high-resolution images by matching their frequency components. Subsequently, we incorporate various visual foundational models as the perceptual guidance for downstream visual tasks, infusing generalizable perceptual features beneficial for detection and segmentation. As a result, our approach gains superior visual results while attaining State-Of-The-Art downstream task performance. Code is available at \url{https://github.com/zirui0625/DifIISR}
\end{abstract}

\section{Introduction}
\label{sec:intro}
The objective of infrared image super-resolution (IISR) is to reconstruct a high-resolution (HR) infrared image from its low-resolution (LR) counterpart~\cite{wang2020deep}. The consistent performance of infrared imaging under challenging conditions allows its application to span various fields~\cite{liu2020bilevel,wang2022unsupervised,wang2023interactively}, such as object detection~\cite{sun2022detfusion,sun2022drone,liu2022target}, semantic segmentation~\cite{li2023text,liu2024promptfusion}, and autonomous driving~\cite{sun2024dynamic, Cao_2023_ICCV}. Despite its great potential, the inherent limitations of infrared cameras — such as high noise levels, reduced spatial resolution, and limited dynamic range — continually affect the quality of infrared images.

Conventionally, CNN-based methods~\cite{zhang2018image, zhang2018residual, zhang2015ccr, luo2020latticenet} address this challenge by mapping interpolated LR images to HR images and then enhancing the details (e.g., SRCNN~\cite{radford2021learning}). Although CNN-based super-resolution methods have significantly advanced this field, they are limited by the perceptual field of local convolution operations. To overcome this, Transformer-based methods~\cite{chen2021pre, cao2021video, zhang2022efficient, zamir2022restormer} model long-range dependencies to capture global context. Liang et al.~\cite{liang2021swinir} proposed SwinIR significantly improving super-resolution performance by integrating CNNs with the Swin Transformer. Lately, Li et al.~\cite{licorple} proposed CoRPLE, which leverages a Contourlet residual framework to restore infrared-specific high-frequency features.

Recently, the diffusion model has introduced a novel paradigm for image super-resolution tasks, offering a fresh approach that goes beyond the CNN- and Transformer-based methods~\cite{cui2024raw}, leveraging its capacity to learn implicit priors of the underlying data distribution~\cite{song2020score}. Yue et al. proposed ResShift~\cite{yue2024resshift}, which applies an iterative sampling procedure to shift the residual between the LR and the desired HR image during inference. Unlike other diffusion models, Wang et al.~\cite{wang2024sinsr} accelerate the diffusion-based SR model to a single inference step while maintaining satisfactory performance. These methods generally achieve visually pleasing results when applied to visible images.

However, existing methods often fail to extend effectively to infrared imaging, particularly in downstream tasks such as infrared image object detection and semantic segmentation. A common approach for task-oriented infrared image super-resolution is to adapt an RGB super-resolution model to infrared data, and then connect it to a downstream detection or segmentation module. Unfortunately, this approach faces two significant challenges: 1) \textbf{Ignoring the unique modal characteristics of infrared imaging}, which include distinct thermal spectrum distributions. Infrared image reconstruction quality is particularly sensitive to high-frequency components due to longer wavelengths and reduced atmospheric scattering effects. 2) \textbf{Overlooking the machine perception requirements}. While the model may reconstruct visually appealing images, these results are often sub-optimal for specific perceptual tasks. The objectives of visual domain optimization and perceptual domain optimization can differ significantly~\cite{liu2022target}. For instance, diffusion-based super-resolution models typically focus on ``seeking visually appealing" results, often at the expense of structural information of targets and textural details critical for machine vision. Given these limitations, we ask, \textbf{``Why not develop a super-resolution model that reconstructs infrared images to be both visually appealing and perceptually salient?"}

To this end, as shown in figure~\ref{fig:teaser}, we propose a task-oriented infrared image super-resolution method that optimizes the diffusion process through gradient-based guidance. Specifically, we inject the gradient of a designed prior loss into the noise estimation at each training step, refining the model’s performance across iterations. Our guidance consists of two components. First, to ensure visual consistency, we introduce visual guidance via infrared thermal spectral distribution modulation, which ensures the reconstructed images align with high-resolution counterparts by preserving their spectral characteristics.  Second, we integrate perceptual guidance by leveraging powerful pre-trained vision models, such as VGG~\cite{simonyan2014very} and SAM~\cite{kirillov2023segment}, to infuse the diffusion process with generalized perceptual features.  Extensive experiments demonstrate that our proposed method excels in both visual quality and downstream task performance. Our contributions can be summarized as follows:

\begin{itemize}
\item We propose a solution for infrared image super-resolution by integrating gradient-based priors into the noise during diffusion, enabling task-based guidance in sampling, and achieving simultaneous optimization in both visual and perceptual-specific domains. 
\item We introduce a thermal spectrum distribution regulation to preserve the visual fidelity of infrared images, guiding the diffusion process to learn the unique infrared image frequency distribution.
\item We propose perceptual guidance for the diffusion process, incorporating generalizable perceptual features from foundational models for visual tasks. This notably enhances performance in detection and segmentation.
\end{itemize}

\section{Related work}
\label{sec:formatting}

\subsection{Image Super-Resolusion}

Since the pioneering work of SRCNN~\cite{dong2014learning} was proposed, deep learning has gradually become the mainstream approach for image super-resolution (SR). The initial works~\cite{dong2014learning, kim2016accurate, zhang2018residual, li2018multi} mainly focused on utilizing convolutional neural networks (CNNs)~\cite{cui2022you} for image super-resolution tasks and optimizing the network by minimizing the mean square error (MSE) between the super-resolved image (SR) and their corresponding high-resolution (HR) counterparts. Subsequently, GAN-based super-resolution methods were proposed, drawing significant attention. For example, both BSRGAN~\cite{zhang2021designing} and Real-ESRGAN~\cite{wang2021real} employ GANs for super-resolution tasks and introduce training samples with more realistic types of degradations to achieve better results. While these methods improve the quality of the low-resolution images, they often fail to produce stable outcomes, resulting in artifacts in the images. LDL~\cite{liang2022details} and DeSRA~\cite{xie2023desra} attempt to address this issue, but they still struggle to generate images with natural details. Recently, diffusion models have been widely applied to image super-resolution tasks, such as ResShift~\cite{yue2024resshift} and SinSR~\cite{wang2024sinsr}. However, these methods are not designed specifically for the characteristics of infrared images and overlook the requirements of machine perception~\cite{zou2024contourlet}, so they do not perform well in infrared image super-resolution (IISR).


\subsection{Diffusion Methods}
The Diffusion Denoising Probabilistic Model (DDPM)~\cite{ho2020denoising} is a generative model with stability and controllability. Since it was proposed, it has attracted widespread attention. The main focus of the diffusion model is to train a denoising autoencoder, which estimates the reverse process of the Markov diffusion process by predicting the noise. Diffusion models were initially applied to image generation tasks and have been continuously improved in recent years~\cite{songdenoising, austin2021structured, nichol2021improved, song2020score, lu2022dpm}. ControlNet~\cite{zhang2023adding} introduces control conditions into pre-trained diffusion models, expanding the application scope of diffusion models in image generation. DDIM~\cite{songdenoising} proposes a non-Markovian generation method, significantly enhancing the inference speed of diffusion models. Diffusion models have demonstrated exceptional capabilities not only in image generation tasks but also in various other tasks, showing great potential. With the introduction of several related methods~\cite{yue2024resshift, wang2024sinsr, saharia2022image, choi2021ilvr, chung2022come, kawar2022denoising}, diffusion models have also been validated to achieve remarkable results in the field of image super-resolution.


\begin{figure*}[!t]
  \centering
  \includegraphics[width=1\textwidth]{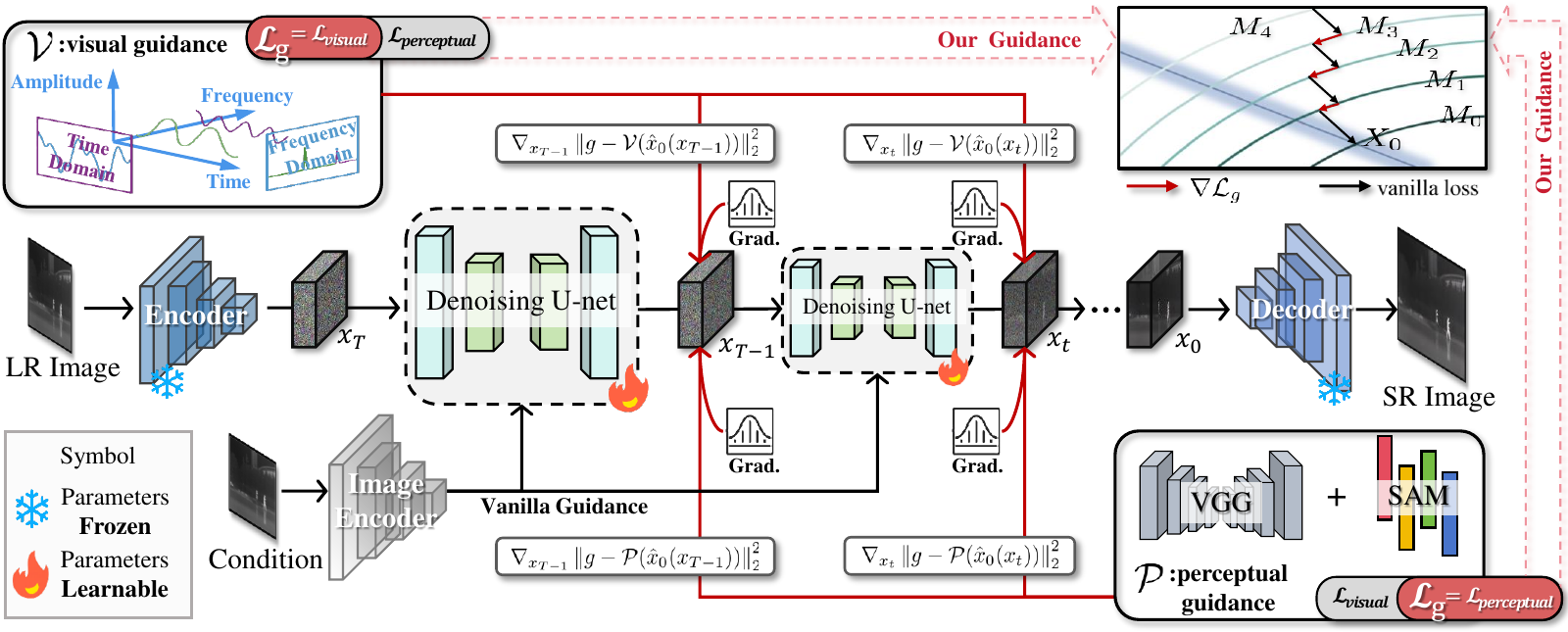}
  \vspace{-0.2in}
  \caption{Overall architecture of our proposed method:  the vanilla super-resolution diffusion process is marked in \textbf{black}, whereas our proposed additional visual and perceptual priors are marked in \textbf{\textcolor{red}{red}}.}
  \label{fig:network}
\end{figure*}

\section{Preliminaries}

\textbf{Diffusion models.} We first introduce the background of Denoising Diffusion Probabilistic Models~\cite{ho2020denoising}. DDPM obtains samples \( x_0 \sim p_{\text{data}}(x) \) from the data distribution. In a diffusion model, noise is gradually added to the sampled \(x_0\) over time steps up to \(T\), eventually resulting in $x_T \sim \mathcal{N}(0, \mathbf{I})$, which can be approximated as a standard Gaussian distribution. This process is also referred to as the forward process of the diffusion model, and it can be represented as:
\begin{equation}
q(x_t \mid x_0) = \mathcal{N}(x_t; \sqrt{\alpha_t} x_0, (1 - \alpha_t) \mathbf{I}),
\end{equation}
where $\alpha_t = \prod_{s=1}^{t} (1 - \beta_s)$, and $\beta_s$ are fixed or learned
variance schedule. After obtaining \( x_T \), the denoising model $\epsilon_{\phi}$ learns to predict the noise \( \epsilon \) added during the forward process, thereby removing the noise from \( x_T \). Specifically, the denoising model $\epsilon_{\phi}$ predicts the noise by optimizing the re-weighted evidence lower bound, which can be written as:
\begin{equation}
\mathcal{L}_{\text{simple}}(\phi) = \mathbb{E}_{x_0, t, \epsilon} \left[ \|\epsilon_{\phi}(x_t, t) - \epsilon\|^2 \right].
\end{equation}
In this formula, $\epsilon_{\phi}(x_t, t)$ represents the noise predicted by the model, and $t$ is randomly sampled from a predefined range of time steps. During the training process, the denoising model $\epsilon_{\phi}$ is optimized by minimizing $\mathcal{L}_{\text{simple}}(\phi)$, ultimately resulting in a model capable of accurately predicting the noise.

After training the denoising model $\epsilon_{\phi}$, we sample $x_T \sim \mathcal{N}(0, \mathbf{I})$ and iteratively refine it using the denoising model. This process is also known as the reverse process, and the specific formula can be represented as:
\begin{equation}
p_{\theta}(x_{t-1} \mid x_t) = \mathcal{N}(x_{t-1}; \mu_{\theta}(x_t, t), \Sigma_{\theta}(x_t, t)),
\end{equation}
where $\mu_{\theta}(x_t, t)$ is the mean function from step $t$ to $t-1$, and $\Sigma_{\theta}(x_t, t)$ is the covariance. Due to the slow sampling process of DDPM, DDIM proposes using a non-Markovian diffusion process, which significantly improves the model's sampling speed. The improved sampling formula can be expressed as:
\begin{equation}
x_{t-1}=\sqrt{\alpha_{t-1}} \hat{x}_0(x_t)+\sqrt{1-\alpha_{t-1}-\sigma_t^2}\epsilon_\phi(x_t, t)+\sigma_t z,
\end{equation}
where $\sigma_t$ is  the variance of the noise and $z$ follows a standard normal distribution. $\hat{x}_0(x_t)$ is the predicted $x_0$ from $x_t$, and the prediction formula is:
\begin{equation}
\hat{x}_0(x_t) = \frac{1}{\sqrt{\alpha_t}} \left( x_t - \sqrt{1 - \alpha_t} \, \epsilon_\phi(x_t, t) \right).
\end{equation}
When $\sigma_t$ equals 0, it is evident that the DDIM sampling process can be regarded as a deterministic process, which allows for quick sampling results from the noise.

\section{Method}
\textbf{Overview.} Our main objective is to address the problem of infrared image super-resolution using a diffusion model enhanced by gradient-based guidance, as shown in figure~\ref{fig:network}. Specifically, inspired by~\cite{chung2023diffusion}, we propose a method that fine-tunes the diffusion model by introducing an additional guidance mechanism. Unlike previous approaches where loss constraints are directly added numerically during training, we compute the gradient of the loss and inject it into the noise predicted at each denoising step. This correction optimizes the denoising process iteratively, refining the model’s output at every stage.  In addition, we incorporate a dual optimization approach combining visual and perceptual aspects to better adapt the diffusion model to the task of infrared image super-resolution.

\subsection{Loss-gradient Guidance}

The reverse process of diffusion models often requires multiple constraints to generate stable, high-quality images. Most methods tend to guide the reverse process by adding weighted constraints to the final loss function. In contrast, our approach addresses this issue from the perspective of posterior sampling. Inspired by~\cite{chung2023diffusion}, we introduce additional priors and compute the gradient of the resulting loss function, injecting the gradient into the noise estimated at each step to better handle the problem of adding constraints during the reverse process of diffusion models.

Generally, the noise predicted by the denoising model at timestep $t$ is often correlated with the score of the denoising model at the current timestep~\cite{song2020score}. Specifically, it can be represented as:
\begin{equation}
\epsilon_{\phi}(x_t, t) = - \sqrt{1 - \alpha_t} \nabla_{x_t} \log p(x_t),
\end{equation}
where $\nabla_{x_t} \log p(x_t)$ is the gradient of $x_t$ with respect to the probability density function $\log p(x_t)$, but now we need to consider not only $\nabla_{x_t} \log p(x_t)$, we also need to incorporate optimization of $g$ during the diffusion model sampling. In our work, \( g \) represents the guidance obtained by feeding \( x_0 \) into $\mathcal{M}$, and $\mathcal{M}$ is a forward operator. The relationship between $g$ and $x_0$ can be expressed as $g = \mathcal{M}(x_0)$. Therefore, the score of the denoising model at timestep $t$ becomes $\nabla_{x_t} \log p(x_t \mid g)$.

$\nabla_{x_t} \log p(x_t \mid g)$ is unknown, and we need to use the known $\nabla_{x_t} \log p(x_t)$ to derive $\nabla_{x_t} \log p(x_t \mid g)$. According to Bayes' theorem, we can write:
\begin{equation}
\nabla_{x_t} \log p(x_t \mid g) = \nabla_{x_t} \log p(x_t) + \nabla_{x_t} \log p(g \mid x_t).
\end{equation}
From this, it can be seen that $\nabla_{x_t} \log p(x_t)$ is known, and the problem changes from calculating $\nabla_{x_t} \log p(x_t \mid g)$ to calculating $\nabla_{x_t} \log p(g \mid x_t)$. Inspired by~\cite{chung2023diffusion}, we can derive the formula:
\begin{equation}
\begin{aligned}
\nabla_{x_t} \log p(g \mid x_t) &\simeq \nabla_{x_t} \log p(g \mid \hat{x}_0(x_t)) \\
&\simeq -\rho \nabla_{x_t} \left\|g - \mathcal{M}(\hat{x}_0(x_t)) \right\|_2^2,
\end{aligned}
\end{equation}
where $\nabla_{x_t} \left\|g - \mathcal{M}(\hat{x}_0(x_t)) \right\|_2^2$ also can be represented as $\nabla \mathcal{L}_{g}$. Therefore, we can express the noise prediction adjusted according to condition $g$ as:
\begin{equation}
\begin{aligned}
\epsilon'_{\phi} &= \epsilon_{\phi}(x_t, t) + \rho \sqrt{1 - \alpha_t} \nabla_{x_t} \left\| g - \mathcal{M}(\hat{x}_0(x_t)) \right\|^2_2  \\
&= \epsilon_{\phi}(x_t, t) + \rho \sqrt{1 - \alpha_t} \nabla \mathcal{L}_{g},
\end{aligned}
\end{equation}
where $\epsilon'_{\phi}$ represents the adjusted noise, obtained by adding the gradient of the guidance loss $\nabla \mathcal{L}_{g}$ to the noise predicted by the original denoising model. 

Thus, by applying gradient guidance to the noise predicted by the diffusion model, we impose constraints on the reverse process of the diffusion model. More detailed derivations and pseudocode of our method can be found in the supplementary materials.

\subsection{Visual-perceptual Dual Optimization}
Now, let's explain the composition of our guidance $\mathcal{L}_{g}$ in detail. Specifically, $\mathcal{L}_{g}$ can be divided into two parts: visual loss $\mathcal{L}_{\text{visual}}$ and perceptual loss $\mathcal{L}_{\text{perceptual}}$.

\noindent\textbf{ $\mathcal{V}$ - visual guidance.}
To guide the diffusion process in reconstructing infrared images towards infrared-specific visual characteristics, we propose \(\mathcal{L}_{\text{visual}}\) to regularize the distribution of high- and low-frequency information as visual guidance $\mathcal{V}$. Here, $\mathcal{V}$  replaces the forward operator $\mathcal{M}$ in equation 8. Given the HR image \( \mathbf{I}_{HR} \) and the super-resolved image \( \mathbf{I}_{SR} \), we first use Fast Fourier Transforms (FFT) to transform their spatial domain representation into the frequency domain, formally:

\begin{equation}
\begin{gathered}
    \hat{\mathbf{I}}_{HR} = \mathcal{F}(\mathbf{I}_{HR}),\, \hat{\mathbf{I}}_{SR} = \mathcal{F}(\mathbf{I}_{SR}),\\
    \mathcal{F}(u, v) = \sum_{x=0}^{H-1} \sum_{y=0}^{W-1} I(x, y) \cdot e^{-i \frac{2\pi}{H} ux} \cdot e^{-i \frac{2\pi}{W} vy}.
\end{gathered}
\end{equation}

\noindent where \( \mathcal{F}(u, v) \) denotes the FFT of the image at frequency coordinates \( (u, v) \), and \( \hat{\mathbf{I}}_{HR} \) is the transformed HR images.
In the frequency domain, we first shift the zero-frequency component, which represents the mean intensity of the image, to the center of the spectrum for both HR and SR images, yielding \( \hat{\mathbf{I}}_{HR}^{\text{shift}} \) and \( \hat{\mathbf{I}}_{SR}^{\text{shift}} \). Following this, we compute the magnitude spectra \( \mathbf{M}_{HR} \) and \( \mathbf{M}_{SR} \) by applying logarithmic compression to the Fourier-transformed images. This step ensures a balanced consideration of both high-frequency and low-frequency components during comparison. To focus the loss on matching the frequency distribution patterns rather than absolute intensity differences, we normalize the magnitude spectra to have zero mean and unit variance, resulting in the normalized spectra \( \mathbf{M}_{HR}^{\text{norm}} \) and \( \mathbf{M}_{SR}^{\text{norm}} \). Finally, the Visual Loss \(\mathcal{L}_{\text{visual}}\) is computed as the mean squared error between the normalized magnitude spectra of the HR and SR images:
\vspace{-0.1in}
\begin{equation}
\mathcal{L}_{\text{visual}}= \biggl( \overbrace{N\bigl(\log(1 + |\hat{\mathbf{I}}_{HR}^{\text{shift}}|)\bigr)}^{\mathbf{M}_{HR}^{\text{norm}}} -  \overbrace{N\bigl(\log(1 + |\hat{\mathbf{I}}_{SR}^{\text{shift}}|)\bigr)}^{\mathbf{M}_{SR}^{\text{norm}}} \biggr)^2,
\end{equation}

\noindent where \(N(\cdot) \) represents the normalization operation. Visual Loss plays a critical role in preserving the frequency distribution of the infrared image.

\begin{table*}[t]
\centering
\small
\renewcommand{\arraystretch}{1.1}
\setlength{\tabcolsep}{3.3mm}
\begin{tabular}{lccccccc}
\Xhline{1.1pt}
\multicolumn{2}{c}{Datasets} & \multicolumn{2}{c}{\textbf{Set5}} & \multicolumn{2}{c}{\textbf{Set15}} & \multicolumn{2}{c}{\textbf{Set20}} \\ 
\cmidrule(lr){1-2} \cmidrule(lr){3-4} \cmidrule(lr){5-6} \cmidrule(lr){7-8}
 \multicolumn{2}{c}{Methods} & CLIP-IQA$\uparrow$ & MUSIQ$\uparrow$ & CLIP-IQA$\uparrow$ & MUSIQ$\uparrow$ & CLIP-IQA$\uparrow$ & MUSIQ$\uparrow$ \\
\hline
\rowcolor[gray]{0.9} Low Resolution\footnotemark & - & 0.2167 & 24.609 &  0.2049 & 23.063 & 0.2230 & 22.446\\
\hline
ESRGAN~\cite{wang2018esrgan} & ECCV'18 & 0.2130 & 40.819 & 0.2038 & 40.745 & 0.1804 & 36.654 \\
RealSR-JPEG~\cite{ji2020real} & CVPR'20 & 0.3615 & 48.419 & 0.3573 & 49.225 & 0.3277 & 47.213 \\
BSRGAN~\cite{zhang2021designing} & CVPR'21 & 0.3290 & 53.119 & 0.3194 & 52.644 & 0.3301 & 51.917 \\
SwinIR~\cite{liang2021swinir} & CVPR'21 & 0.2160 & 37.156 & 0.2230 & 37.970 & 0.2258 & 34.919 \\
RealESRGAN~\cite{wang2021real} & ICCV'21 & 0.2780 & 54.306 & 0.2424 & 53.163 & 0.2523 & 51.647 \\
HAT~\cite{chen2023activating} & CVPR'23 & 0.2298 & 38.050 & 0.2377 & 39.743 & 0.2466 & 35.633 \\
DAT~\cite{chen2023dual} & ICCV'23 & 0.2297 & 37.538 & 0.2410 & 39.419 & 0.2518 & 35.750 \\
ResShift~\cite{yue2024resshift} & NeurIPS'23 & 0.4701 & 50.769 & 0.4428 & 52.871 & 0.4082 & 51.244 \\
CoRPLE~\cite{licorple} & ECCV'24 & 0.2339 & 36.281 & 0.2281 & 36.458 & 0.2281 & 34.270\\
SinSR~\cite{wang2024sinsr} & CVPR'24 & \underline{0.5877} & \underline{54.355} & \underline{0.5762} & \underline{54.106} & \underline{0.5357} & \underline{53.187}\\
Bi-DiffSR~\cite{chen2024binarized} & NeurIPS'24 & 0.3151 & 35.356 & 0.2758 & 36.102 &  0.2674 & 36.537\\
\hline
\rowcolor[gray]{0.9} DifIISR & Ours & \textbf{0.6144} & \textbf{55.194} & \textbf{0.5906} & \textbf{54.504} & \textbf{0.5484} & \textbf{53.636} \\
\rowcolor[gray]{0.9} High Resolution & - & 0.2200 & 34.066 &  0.2161 & 34.410 & 0.2139 & 32.024\\
\Xhline{1.1pt}
\end{tabular}
\caption{No-reference Metrics Comparison of infrared image super-resolution with SOTA methods on \(\text{M}^3\text{FD}\) datasets.}
\label{Table: compare_m3fd}
\end{table*}

\begin{table*}[h]
\small
\centering
\renewcommand{\arraystretch}{1.1}
\setlength{\tabcolsep}{1.3mm}
\begin{tabular}{l|cccc|cccc|cccc}
\Xhline{1.1pt}
& \multicolumn{12}{c}{Datasets} \\
\cline{2-13}
\multirow{1}{*}{Metrics}
& \multicolumn{4}{c}{\textbf{Set5}} & \multicolumn{4}{c}{\textbf{Set15}} & \multicolumn{4}{c}{\textbf{Set20}} \\
\cline{2-13}
 & ResShift & SinSR & Bi-DiffSR & DifIISR & ResShift & SinSR & Bi-DiffSR & DifIISR & ResShift & SinSR & Bi-DiffSR & DifIISR  \\
\hline 
    PSNR$\uparrow$ & 30.101 & 31.645 & \underline{32.022} & \textbf{32.279} & 30.283 & 31.988 & \underline{32.145} & \textbf{32.351} & 30.976 & 33.438 & \underline{33.447} & \textbf{33.451}\\
SSIM$\uparrow$  & 0.8329 & 0.8481 & \underline{0.8579}& \textbf{0.8637} & 0.8228 & 0.8426 & \underline{0.8471} & \textbf{0.8578} & 0.8446 & 0.8853 & \underline{0.8874} & \textbf{0.8941} \\
LPIPS$\downarrow$& 0.3179 & \underline{0.2737} & 0.2816 & \textbf{0.2704}& 0.3537 & \textbf{0.2817} & 0.2924 & \underline{0.2845} & 0.3507 & \textbf{0.2549} & 0.2820 & \underline{0.2735} \\
\Xhline{1.1pt}
\end{tabular}
\caption{Reference-based Metrics Comparison with diffusion-based methods on \(\text{M}^3\text{FD}\) datasets.}
\label{Table: diffusion-based methods}
\end{table*}

\noindent\textbf{ $\mathcal{P}$ - perceptual guidance.}
To regularize the diffusion process to better align with machine perception, we adopt \(\mathcal{L}_{\text{perceptual}}\) that consists of the VGG Loss \(\mathcal{L}_{\text{VGG}}\) and the Segmentation Loss \(\mathcal{L}_{\text{seg}}\) as perceptual guidance $\mathcal{P}$. Here, $\mathcal{P}$ replaces the forward operator $\mathcal{M}$ in equation 8. Given the HR image \( \mathbf{I}_{HR} \) and the super-resolved image \( \mathbf{I}_{SR} \), the VGG Loss is computed by mean squared error between the extracted features of the HR and SR images from a pre-trained deep neural network. This guides the model to capture nuanced aspects of images, including textures, edges, and shapes, which are crucial for preserving visual fidelity. To enhance the semantic fidelity of the reconstructed images, we regulate the diffusion process using the Segment Anything Model (SAM)~\cite{kirillov2023segment} and propose the Segmentation Loss \(\mathcal{L}_{\text{seg}}\). Given the HR image \( \mathbf{I}_{HR} \) and the super-resolved image \( \mathbf{I}_{SR} \), we use a locked SAM to segment the masks \( \mathbf{S}_{HR} \) and \( \mathbf{S}_{SR} \) for \( \mathbf{I}_{HR} \) and \( \mathbf{I}_{SR} \), respectively. The Segmentation Loss \(\mathcal{L}_{\text{seg}}\) is then computed by mean squared error between \( \mathbf{S}_{HR} \) and \( \mathbf{S}_{SR} \), providing effective high-level supervision for the reconstructed images.

The Perceptual Loss \(\mathcal{L}_{\text{perceptual}}\) is computed by integrating the VGG-based and segmentation-based losses, as:
\vspace{-0.1in}
\begin{equation}
\mathcal{L}_{\text{perceptual}} = \overbrace{\left\| \phi_l(\mathbf{I}_{HR}) - \phi_l(\mathbf{I}_{SR}) \right\|_2^2}^{\mathcal{L}_{\text{VGG}}} + \overbrace{\left\|\mathbf{S}_{HR} - \mathbf{S}_{SR} \right\|_2^2}^{\mathcal{L}_\text{seg}},
\end{equation}

\noindent where \( \phi_l(\cdot) \) represents the feature map extracted from the \( l \)-th layer of a pre-trained deep neural network (in our experiment, VGG-16). 

The incorporation of visual and perceptual guidance refines each iteration of the diffusion, facilitating a more optimized denoising procedure. This not only improves visual fidelity but also enhances perceptual performance.


\begin{figure*}[!t]
 \centering
 \includegraphics[width=1\textwidth]{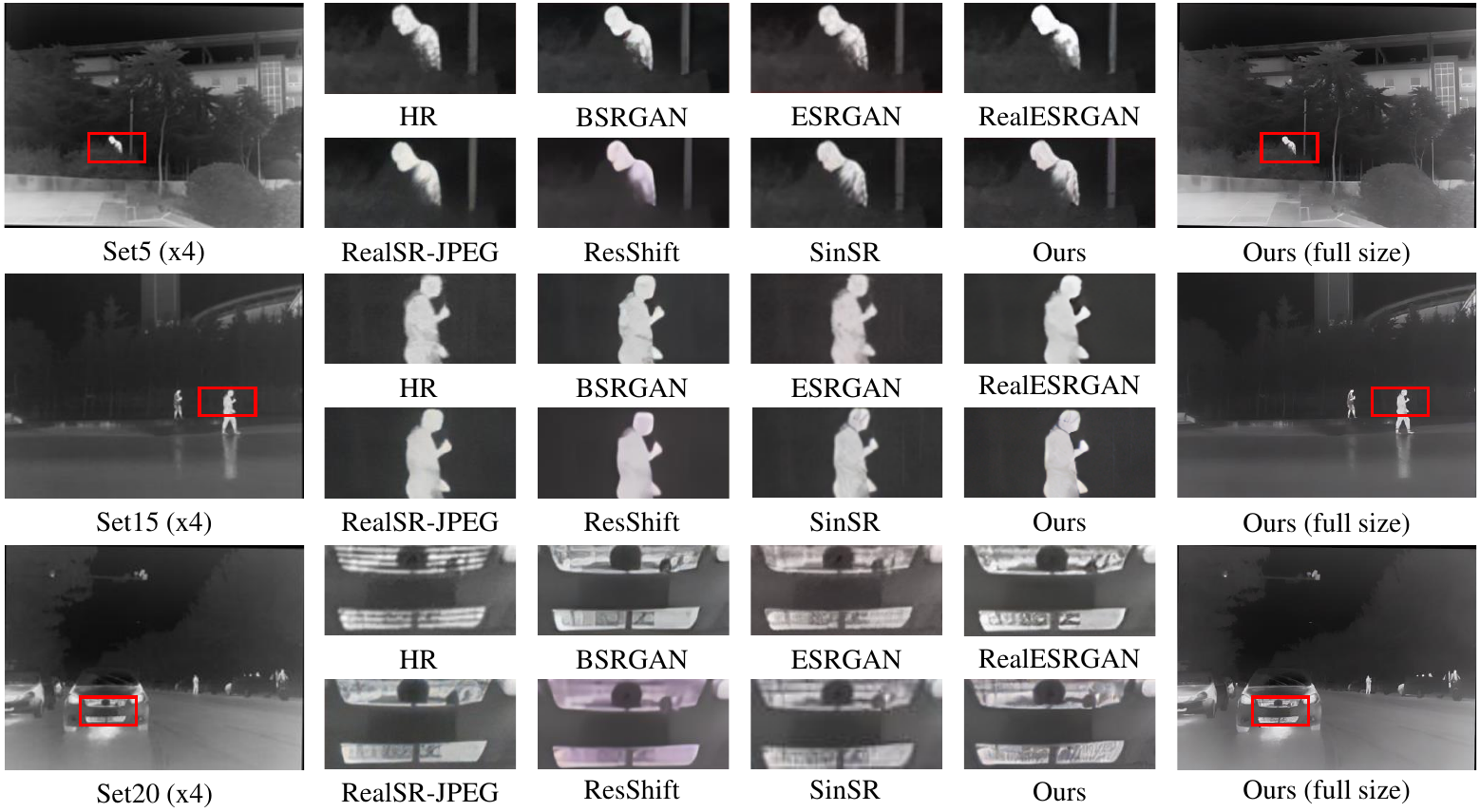}
    \caption{Visual comparison of infrared image super-resolution with SOTA methods on \(\text{M}^3\text{FD}\) datasets.}
    \label{Exp: compare_m3fd}
\end{figure*}

\section{Experiments}

\subsection{Experimental Settings}
\textbf{Dataset and evaluation metrics.}
To ensure the fairness of the experiment, we used the same training~\cite{liu2022target} and test sets~\cite{liu2022target, xu2020u2fusion, toet2017tno}, as CoRPLE~\cite{licorple}.
We use the infrared image dataset $\text{M}^3\text{FD}$~\cite{liu2022target} to train the model and evaluate its performance using three datasets: $\text{M}^3\text{FD}$~\cite{liu2022target}, RoadScene~\cite{xu2020u2fusion}, and TNO~\cite{toet2017tno}.
We adopt five metrics to evaluate the performance of our model quantitatively: CLIPIQA~\cite{wang2023exploring}, MUSIQ~\cite{ke2021musiq}, PSNR, LPIPS~\cite{zhang2018unreasonable}, and SSIM~\cite{wang2004image}. Among them, CLIPIQA and MUSIQ are no-reference metrics. CLIPIQA leverages the CLIP model~\cite{radford2021learning} to assess image quality, while MUSIQ uses a multi-scale feature extraction approach for quality evaluation. We mainly rely on CLIPIQA and MUSIQ as evaluation metrics to compare the performance of different methods.

\noindent\textbf{Implementation Details.} Our network was trained on a GeForce RTX 4090 GPU. Our backbone model and specific experimental parameter settings largely follow ResShift~\cite{yue2024resshift}. Notably, ResShift uses the residual between high-resolution (HR) and low-resolution images (LR) as the noise for the diffusion model, meaning that we can effectively apply gradient guidance on the residual between HR and LR images. During training, our approach differs from ResShift in that we initially perform 200K iterations on a new training set to enable the model to develop basic infrared image super-resolution capabilities. Subsequently, we incorporate conditional (visual and perceptual) guidance into the model and conduct an additional 50K training iterations to achieve improved results.

\subsection{Experiments on Infrared SR}
We perform a comprehensive comparison of our approach with eleven SOTA methods, including ESRGAN~\cite{wang2018esrgan}, RealSR-JPEG~\cite{ji2020real}, BSRGAN~\cite{zhang2021designing}, SwinIR~\cite{liang2021swinir}, RealESRGAN~\cite{wang2021real}, HAT~\cite{chen2023activating}, DAT~\cite{chen2023dual}, ResShift~\cite{yue2024resshift}, CoPRLE~\cite{licorple}, Bi-DiffSR~\cite{chen2024binarized} and SinSR~\cite{wang2024sinsr}. Table \ref{Table: compare_m3fd}, \ref{Table: diffusion-based methods} presents our quantitative comparison results compared with the above methods and Figure \ref{Exp: compare_m3fd} presents our qualitative results.

\noindent\textbf{Quantitative Comparison.}
Table \ref{Table: compare_m3fd} presents a quantitative comparison of CLIPIQA and MUSIQ on the $\text{M}^3\text{FD}$ dataset against various methods. CLIPIQA inherits the powerful representation capabilities of CLIP, demonstrating stable and robust performance in evaluating the perceptual quality of natural images. Our method outperforms other methods on both metrics across all three test sets, indicating that our approach better aligns with the human perceptual system. Additionally, our method achieves superior performance on MUSIQ compared to all other methods, demonstrating that it can also achieve excellent results in multi-scale image quality assessment. 

It is worth noting that we also compared our method against HR images on no-reference metrics. Our method significantly outperforms HR images in no-reference visual quality metrics, demonstrating an enhancement over the HR images. However, this improvement introduces a challenge: in comparison to traditional methods on reference-based metrics such as PSNR, LPIPS, and SSIM, our approach shows less advantage, as our results differ significantly from the HR images. Nevertheless, our method still leads diffusion-based methods on reference-based metrics, as shown in Table \ref{Table: diffusion-based methods}. This demonstrates that our approach leverages the powerful generative capabilities of diffusion to produce high-quality images while also preserving essential detail features from the HR images under both visual and perceptual guidance.

\footnotetext{Evaluate the low-resolution image after enlarging it to match the resolution of the high-resolution image through interpolation.}

\noindent\textbf{Qualitative Results.}
The qualitative results shown in Figure \ref{Exp: compare_m3fd} emphasize the superiority of our method in visual performance compared to other approaches. Additional examples can be found in the supplementary materials. We selected one image from each of the three datasets, \textbf{Set5}, \textbf{Set15}, and \textbf{Set20}, for qualitative analysis to ensure comprehensive evaluation. Our method achieves more natural details in portraits, avoiding color discrepancies and better matching the contours of the true image. For vehicle details, our method accurately reproduces the grille at the front of the vehicle in the true image, whereas other methods tend to blur these details. This demonstrates that our method also has distinct advantages in qualitative results.

\subsection{Ablation Study.}

\noindent\textbf{Experiments on the effectiveness of guidance.}
We conducted ablation experiments to evaluate the effectiveness of visual and perceptual guidance on the infrared super-resolution task, as shown in Table \ref{abla1}. We assessed the super-resolution results under four conditions: without guidance, with only visual guidance, with only perceptual guidance, and with both visual and perceptual guidance. The results show that the infrared image super-resolution performance is best when both guidance are applied.

\noindent\textbf{Experiments on the guidance combinations.}
We conducted ablation experiments on different guidance combinations of various methods, as shown in Table \ref{abla2}. In the perceptual-based setup, which involves using a perceptual loss gradient for guidance, we performed three sets of experiments: (1) without the visual loss, (2) directly adding the visual loss \(\sum{\mathcal{L}}\), and (3) incorporating the gradient of the loss \(\nabla{\mathcal{L}}\) into the noise. The experimental results demonstrate that incorporating the gradient of the loss into the noise yields the best performance. We also conducted experiments in a visual-based setup, the results under the visual-based setup also follow this trend.

\begin{table}[!t]
\centering
\small
\renewcommand{\arraystretch}{1.1}
\setlength{\tabcolsep}{2mm}
\begin{tabular}{cccccc}
\Xhline{1.1pt}
Visual  & Perceptual  & PSNR   & CLIP-IQA & mAP & mIoU \\ \hline
-            & -          & 33.466 & 0.5102   & 31.2   & 40.9    \\
\checkmark   & -          & \underline{34.528} & \underline{0.5365}   & 31.7   & 41.3    \\
-            & \checkmark          & 33.923 & 0.5230    & \underline{32.8}   & \underline{42.2}    \\
\checkmark   & \checkmark          & \textbf{34.575} & \textbf{0.5379}   & \textbf{33.1}   & \textbf{42.4}    \\ \Xhline{1.1pt}
\end{tabular}
\caption{Ablation study on the effectiveness of multiple guidance.}
\label{abla1}
\end{table}

\begin{figure*}[!t]
 \centering
 \includegraphics[width=1\textwidth]{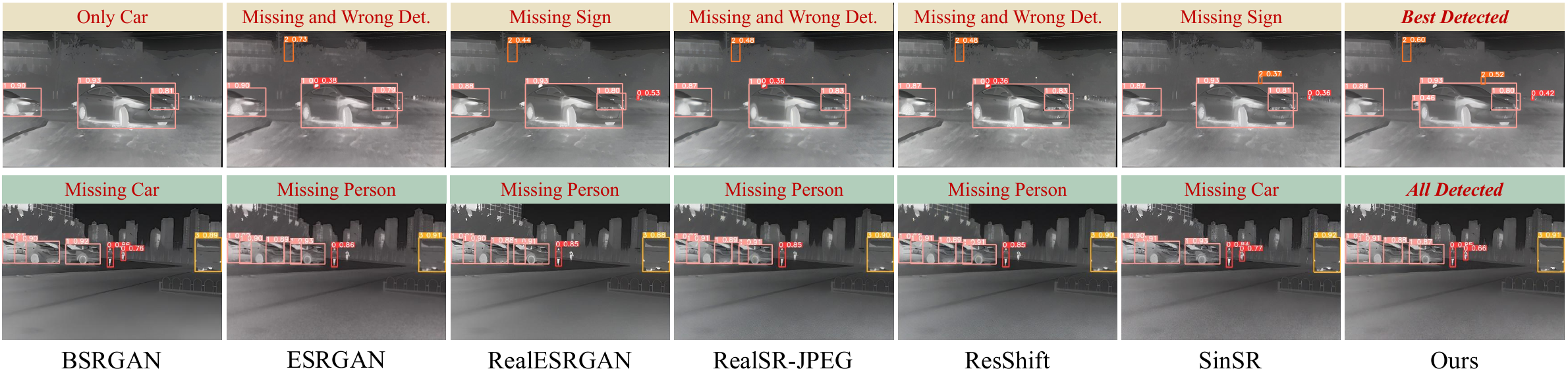}
 \vspace{-0.15in}
    \caption{Detection performance comparison of infrared image super-resolution with SOTA methods on \(\text{M}^3\text{FD}\) datasets.}
    \label{Exp: compare_det}
\end{figure*}

\begin{figure*}[!t]
 \centering
 \includegraphics[width=1\textwidth]{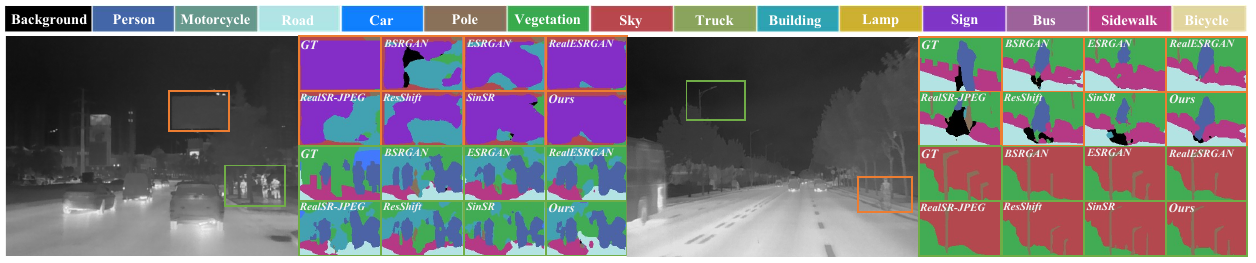}
  \vspace{-0.15in}
    \caption{Segmentation performance comparison of infrared image super-resolution with SOTA methods on FMB datasets.}
    \label{Exp: compare_seg}
\end{figure*}

\begin{figure}[!t]
 \centering
 \includegraphics[width=0.47\textwidth]{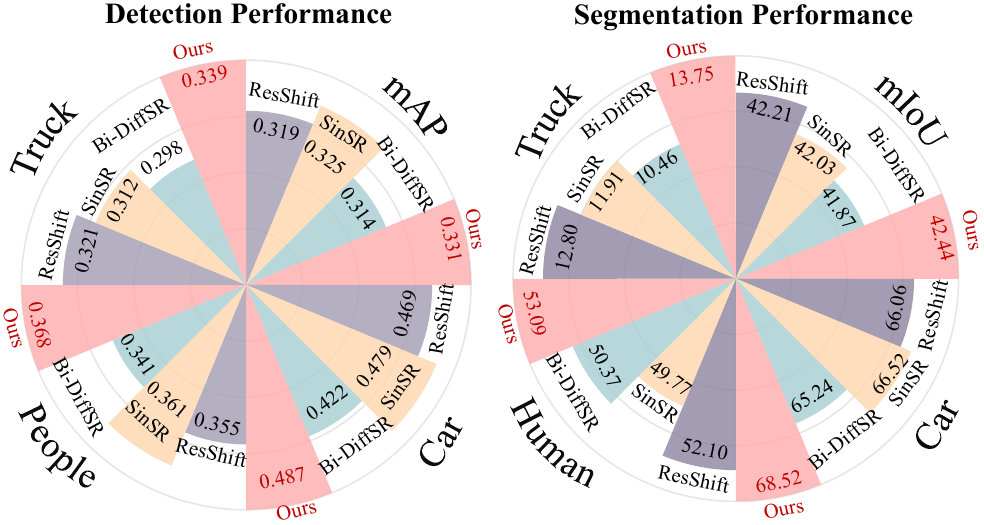}
    \caption{Quantitative comparison of detection and segmentation results with SOTA methods.}
    \label{Exp: compare_metric}
\end{figure}

\subsection{Experiments on Infrared Object Detection}

\textbf{Setup.}  We employ YOLOv5-s for infrared image object detection, fine-tuning it specifically on the \(\text{M}^3\text{FD}\) dataset. The primary evaluation metric is the mean Average Precision (mAP) at varying IoU thresholds (mAP@.5:.95). The model is fine-tuned with a batch size of 16, using the SGD optimizer with learning rate of 0.01.

\noindent\textbf{Quantitative Comparison.} The left section of Figure \ref{Exp: compare_metric} presents a quantitative comparison of detection results across SOTA methods. In the top-right quadrant of the plot, the overall mAP of each model is displayed, while the other three quadrants represent the performance across individual categories. Our model consistently outperforms all other models in each detection category, demonstrating its superior ability in object detection tasks. Notably, in the truck detection category, our model achieves a 5.6\% improvement over the best-competing method, underscoring its robustness in identifying challenging classes.

\begin{table}[!t]
\centering
\small
\renewcommand{\arraystretch}{1.1}
\setlength{\tabcolsep}{1.4mm}
\begin{tabular}{lccccc}
\Xhline{1.1pt}
 &Guide           & PSNR   & CLIP-IQA & mAP & mIoU \\ \hline
Perceptual Base    & -    & 33.923 & 0.5230    & \underline{32.8}   & \underline{42.2}   \\
\quad + Visual       & $\sum\mathcal{L}$ & \underline{34.061} & \underline{0.5342}   & 32.5   & 42.0   \\
\quad + Visual       & $\nabla{\mathcal{L}}$    & \textbf{34.575} & \textbf{0.5379}   & \textbf{33.1}   & \textbf{42.4}  \\ \hline
Visual Base  & -    & 34.528 & 0.5365   & 31.7   & 41.3    \\
\quad + Task       & $\sum\mathcal{L}$    & \underline{34.561} & \underline{0.5371}   & \underline{32.8} & \underline{41.9}   \\ 
\quad + Task         & $\nabla{\mathcal{L}}$ & \textbf{34.575} & \textbf{0.5379}   & \textbf{33.1}   & \textbf{42.4}  \\ \Xhline{1.1pt}
\end{tabular}
\caption{Ablation study for different guidance combinations.}
\label{abla2}
\end{table}

\noindent\textbf{Qualitative Comparison.} The qualitative results in Figure \ref{Exp: compare_det} demonstrate the superiority of our method in object detection. Other methods frequently miss at least one label or make errors. For example, in the first row, some methods fail to detect the person on the right side of the image, with none capable of detecting both signs above simultaneously. In the second row, certain methods miss the people farthest away, and others are unable to recognize the partially obstructed car. Only our method consistently achieves the best detection prediction results.

\subsection{Experiments on Infrared Image Segmentation}

\noindent\textbf{Setup.} We perform semantic segmentation on the FMB dataset~\cite{liu2023segmif}. The SegFormer-b1 model~\cite{xie2021segformer} is used as the backbone, with intersection-over-union (IoU) as the primary evaluation metric. Supervised by cross-entropy loss, the model is trained using the AdamW optimizer, with a learning rate of 6e-05 and a weight decay of 0.01. Training spans 25,000 iterations with a batch size of 8.

\noindent\textbf{Quantitative Comparison.} The right section of Figure \ref{Exp: compare_metric} presents a quantitative comparison of semantic segmentation results. The top-right quadrant of the circle represents the comparison of mIoU, while the remaining three quadrants depict the performance of other models across the three primary segmentation classes. Overall, our model achieves the best results in each category. Notably, it achieves the highest improvement in truck, with an improvement of 7.4\%. It also improves by 5.4\% and 3.0\% in car and human, respectively.

\noindent\textbf{Qualitative Comparison.} The figure \ref{Exp: compare_seg} presents a qualitative comparison of segmentation results from various SOTA methods. These results reveal that other methods often fail to segment complete objects, or they struggle with segmenting all relevant elements. For example, in other models, only part of the sign occurs, leaving parts of it undetected. While RealESRGAN shows some improvement, it still falls short of our method. Similarly, in the right image, other models fail to recognize the farthest poles and cannot fully capture the shapes of the people.

\section{Conclusion}
In this paper, we propose a task-oriented infrared image super-resolution diffusion model, namely DifIISR. Specifically, we introduce infrared thermal spectral distribution modulation as visual guidance to ensure consistency with high-resolution images by matching frequency components. In addition, we incorporate foundational vision models to provide perception guidance, which enhances detection and segmentation performance. With the above guidance, our method further optimizes each iteration of the standard diffusion process, refining the model at each denoising step and achieving superior visual and perceptual performance. 
{
    \small
    \bibliographystyle{ieeenat_fullname}
    \bibliography{main}
}

\end{document}